\documentclass[conference]{IEEEtran}
\IEEEoverridecommandlockouts

\usepackage{cite}
\usepackage{amsmath,amssymb,amsfonts}
\usepackage{algorithmic}
\usepackage[linesnumbered,ruled]{algorithm2e}
\usepackage{graphicx}
\usepackage{textcomp}
\usepackage[table]{xcolor}
\usepackage{amsthm}

\usepackage[T1]{fontenc}
\bstctlcite{IEEEexample:BSTcontrol}
\usepackage{xcolor}
\usepackage[colorlinks=true, linkcolor=black]{hyperref}
\usepackage{booktabs}
\usepackage{makecell}

\usepackage{multirow}

\def\BibTeX{{\rm B\kern-.05em{\sc i\kern-.025em b}\kern-.08em
    T\kern-.1667em\lower.7ex\hbox{E}\kern-.125emX}}

\makeatletter
\newcommand{\linebreakand}{
  \end{@IEEEauthorhalign}
  \hfill\mbox{}\par
  \mbox{}\hfill\begin{@IEEEauthorhalign}
}
\makeatother

\begin{document}

\title{Detecting Autism Spectrum Disorder with Deep Eye Movement Features

}

\author{\IEEEauthorblockN{1\textsuperscript{st} Zhanpei Huang}
\IEEEauthorblockA{\textit{School of Computer Science and Technology} \\
\textit{Guangdong University of Technology}\\
Guangzhou, China \\
2112405010@mail2.gdut.edu.cn}
\and
\IEEEauthorblockN{2\textsuperscript{nd} Taochen Chen}
\IEEEauthorblockA{\textit{School of Computer Science and Technology} \\
\textit{Guangdong University of Technology}\\
Guangzhou, China \\
chentaochen1@mails.gdut.edu.cn}

\linebreakand
\IEEEauthorblockN{3\textsuperscript{rd} Fangqing Gu}
\IEEEauthorblockA{\textit{School of Mathematics and Statistics} \\
\textit{Guangdong University of Technology}\\
Guangzhou, China \\
fqgu@gdut.edu.cn}
\and
\IEEEauthorblockN{4\textsuperscript{th} Yiqun Zhang\IEEEauthorrefmark{1}\thanks{\IEEEauthorrefmark{1} Corresponding author: Yiqun Zhang (yqzhang@gdut.edu.cn) }}
\IEEEauthorblockA{\textit{School of Computer Science and Technology} \\
\textit{Guangdong University of Technology}\\
Guangzhou, China \\
yqzhang@gdut.edu.cn}
}

\maketitle

\begin{abstract}
Autism Spectrum Disorder (ASD) is a neurodevelopmental disorder characterized by deficits in social communication and behavioral patterns. Eye movement data offers a non-invasive diagnostic tool for ASD detection, as it is inherently discrete and exhibits short-term temporal dependencies, reflecting localized gaze focus between fixation points. These characteristics enable the data to provide deeper insights into subtle behavioral markers, distinguishing ASD-related patterns from typical development. Eye movement signals mainly contain short-term and localized dependencies. However, despite the widespread application of stacked attention layers in Transformer-based models for capturing long-range dependencies, our experimental results indicate that this approach yields only limited benefits when applied to eye movement data. This may be because discrete fixation points and short-term dependencies in gaze focus reduce the utility of global attention mechanisms, making them less efficient than architectures focusing on local temporal patterns. To efficiently capture subtle and complex eye movement patterns, distinguishing ASD from typically developing (TD) individuals, a discrete short-term sequential (DSTS) modeling framework is designed with Class-aware Representation and Imbalance-aware Mechanisms. Through extensive experiments on several eye movement datasets, DSTS outperforms both traditional machine learning techniques and more sophisticated deep learning models.

\end{abstract}


\section{Introduction}

\renewcommand{\thefootnote}{}
\footnotetext{This work was supported in part by the National Natural Science Foundation of China under Grant 62476063 and the Natural Science Foundation of Guangdong Province under Grant 2025A1515011293.}
\renewcommand{\thefootnote}{\arabic{footnote}}

Autism Spectrum Disorder (ASD) is a complex and heterogeneous neurodevelopmental disorder~\cite{lord2020autism}, marked by deficits in social communication, restricted and repetitive behaviors, and sensory processing challenges. Early diagnosis is critical, as timely intervention during crucial developmental windows can significantly improve long-term developmental outcomes. Current diagnostic methods for ASD, however, often rely on behavioral observations and subjective assessments, such as the Autism Diagnostic Observation Schedule (ADOS)~\cite{lord2000autism} and the Autism Diagnostic Interview-Revised (ADI-R)~\cite{rutter2003autism}, which require highly trained clinicians. These methods can be time-consuming, resource-intensive, and often exhibit inter-rater variability, making them difficult to implement consistently across diverse clinical environments.

\begin{figure}
    \centering
    \includegraphics[width=1\linewidth]{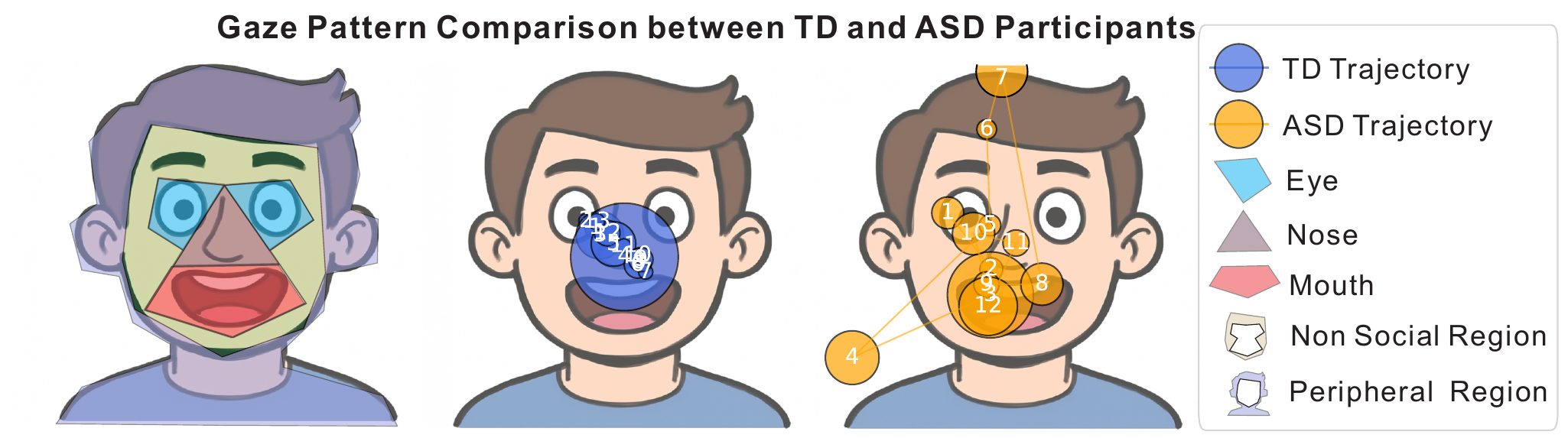}
    \caption{Gaze Pattern Comparison between TD and ASD Participants. The numbers denote the sequential order of fixations, and the circle size corresponds to the duration of each fixation.}
    \label{fig:gaze_plot}
\end{figure}

In recent years, eye movement data has shown promise as a non-invasive diagnostic tool for identifying ASD-related behavioral markers~\cite{billeci2016disentangling}. Eye-tracking technology captures detailed, real-time data on an individual’s gaze patterns and visual focus, providing insights into cognitive processing and social gaze. Children with ASD exhibit distinctive eye movement patterns, such as reduced fixation on faces, atypical scan path patterns, and deviations in saccadic eye movements, reflecting underlying social and cognitive processing differences.
When exposed to stimuli (see Fig.~\ref{fig:gaze_plot}), the eye movement patterns of typically developing (TD) children and children with autism spectrum disorder (ASD) differ significantly. TD children typically focus more on social regions, such as the mouth and eyes, especially under social stimuli. In contrast, children with ASD show reduced gaze toward these areas and tend to focus more on non-social stimuli.
These patterns are often subtle and not easily discernible using traditional clinical methods. Thus, effective machine learning techniques are required to process and interpret this complex, discrete time-series data to distinguish between ASD and TD children.

Traditional approaches to analyzing eye movement data have often relied on manually engineered features~\cite{raj2020analysis, kong2022different}. These features are then used with classifiers like Support Vector Machines (SVM) for pattern recognition~\cite{kang2020identification} and linear discriminant analysis~\cite{constantino2017infant}. Support Vector Regression (SVR) has also been applied to predict Alzheimer’s disease, demonstrating the potential of eye movement biomarkers in neurological disorder assessment~\cite{xu2024portable}. The main advantage of these methods lies in their reliance on expert knowledge, which allows for a focused extraction of eye movement characteristics known to correlate with behavioral markers.

Deep learning models have been widely applied in disease detection and medical signal analysis, as demonstrated by DE3S~\cite{TaoXie2025}, adaptive CNN ensemble methods for biomarker identification~\cite{zeng2021discovery}, and vision-transformer-based frameworks such as CmdVIT~\cite{ye2025cmdvit} for mental disorder recognition. As for eye movement analysis, such as Long Short-Term Memory (LSTM)~\cite{zhou2024gaze}, has been employed for gaze pattern analysis in ASD classification.
Transformer-based architectures~\cite{transformerdisease} have also gained attention for their ability to capture long-term dependencies within sequential data. Initially applied to tasks such as flu outbreak prediction, Transformer models employ self-attention mechanisms to model complex relationships~\cite{ye2024mad,ye2025mssnet}. Variants such as Informer~\cite{zhou2021informer}, Crossformer~\cite{zhang2023crossformer} further enhance efficiency and spatio-temporal representation, while MPTSNet~\cite{mu2025mptsnet}, and EmMixFormer~\cite{qin2025emmixformer} integrate convolutional and multi-domain attention mechanisms to improve feature learning for time-series data.
However, most of these Transformer-based architectures are designed to exploit long-range temporal dependencies and may not be optimally aligned with the discrete, short-term nature of eye movement signals.

Attention mechanisms are designed to capture long-range dependencies by dynamically weighting temporal relationships across a sequence. However, as shown in Fig.~\ref{fig:gaze_plot}, eye movement data exhibit distinct temporal characteristics. Each sequence comprises discrete fixation points with varying durations under continuously changing stimuli. Over long timescales, these fixation sequences do not follow stable dependency patterns, as gaze behavior rapidly adapts to new visual information. In contrast, adjacent fixation points are strongly correlated, reflecting short-term perceptual continuity and localized gaze shifts. In Transformer architectures, the global self-attention mechanism considers pairwise interactions across the entire sequence, which may inadvertently aggregate weakly related or unrelated gaze tokens and thereby introduce additional noise. This property suggests that long-term dependency modeling is less effective for eye-tracking data, while models emphasizing local temporal relationships are more aligned with the intrinsic dynamics of gaze behavior.

To address the discrete structure and short-term dependency characteristics of the eye movement data, a discrete short-term sequential (DSTS) modeling framework is designed that captures localized temporal patterns within fixation sequences. To align with this property, the framework adopts convolutional architectures.
Owing to their localized receptive fields, these architectures naturally emphasize short-term temporal dependencies and capture fine-grained local variations in the fixation sequence.
Furthermore, the model incorporates mechanisms that promote class-aware representation learning and mitigate the effects of data imbalance, enabling it to learn more discriminative and stable gaze representations across heterogeneous subjects.

Our key contributions are as follows:
\begin{itemize}
    \item To address the challenges posed by the discrete nature and short-term dependencies of eye movement data, this work proposes a framework that incorporates class-aware representation learning and imbalance-aware mechanisms, effectively capturing localized temporal patterns and improving discriminative gaze representations across heterogeneous subjects.
    \item This work reveals that stacking attention mechanisms is ineffective for the discrete short-term dependencies in eye movement data, a novel finding that provides inspiration and a new paradigm for addressing similar challenges in other application domains.

\end{itemize}

\section{Related Work} \label{sec:related_work}

In this section, we provide an overview of the existing literature relevant to our study, focusing on time-series analysis and eye movement recognition for disease diagnosis.

\subsection{Time series Analysis}

Time-series data is a common feature in many real-world applications, including eye movement analysis. Recent advances in time-series modeling have leveraged deep learning techniques to capture long-term dependencies across sequences.  One notable approach is the Informer~\cite{zhou2021informer}, which integrates Transformer in time-series data to capture temporal latent features. Another method, Crossformer~\cite{zhang2023crossformer}, focuses on modeling both spatial interactions and temporal dependencies using cross-attention mechanisms. 

Recently, iTransformer uses an iterative Transformer architecture for sequence modeling~\cite{liu2023itransformer}. iTransformer is specifically designed to improve the efficiency of the standard Transformer, particularly when dealing with long sequences. EmMixformer~\cite{qin2025emmixformer} leverages multiple techniques, including time-domain features, Fourierformer, and attLSTM, to capture both short-term and long-term dependencies, as well as frequency-domain characteristics, for learning temporal data features. On the other hand, MPTSNet employs a combination of CNN and attention mechanisms. DE3S~\cite{TaoXie2025} proposes a dual-enhanced soft–sparse–shape learning framework designed for early medical classification. EMTC~\cite{tan2025mask} introduces an evolving masking–based representation learning strategy to address redundancy in multivariate time-series data. These methods all incorporate the Transformer architecture to enhance their feature learning capabilities. 
However, existing models mainly focus on continuous physiological time-series or global dependencies, leaving the discrete and short-term fixation characteristics of eye movements insufficiently explored.

\subsection{Eye Movement for Disease Recognition}

Eye movement analysis has also been extensively applied to the recognition of diseases, such as Parkinson's disease and Autism Spectrum Disorder (ASD). Traditional methods in this domain often involve classifiers like Support Vector Machines (SVM)~\cite{kong2022different}. These models aim to capture patterns in gaze data that are indicative of specific disease-related behaviors.

In the context of Parkinson's disease, eye movement characteristics, including scan paths and fixation duration, are valuable for distinguishing patients from healthy individuals. However, for a more advanced approach, Detach-ROCKET~\cite{uribarri2024detach} has shown promise. Detach-ROCKET builds on the random convolutional kernel method, ROCKET, and introduces a feature pruning technique known as Sequential Feature Detachment (SFD). This method identifies the most informative features while reducing the computational complexity, making it especially useful in time-series classification tasks such as Parkinson's disease detection using eye-tracking data~\cite{uribarri2023deep}.

Recent advancements have seen the use of deep learning models, including LSTM~\cite{zhou2024gaze}, to capture and classify gaze behavior in individuals with ASD, particularly focusing on deficits in social attention, such as reduced fixation on faces and atypical gaze shifts during social tasks. 
However, these models typically assume smooth temporal evolution and do not explicitly account for the discrete fixation structure, abrupt saccadic transitions, and short-term dependencies that dominate eye-movement sequences.

\section{Proposed Method}\label{sec_pm}

\begin{figure}
    \centering
    \includegraphics[width=1\linewidth]{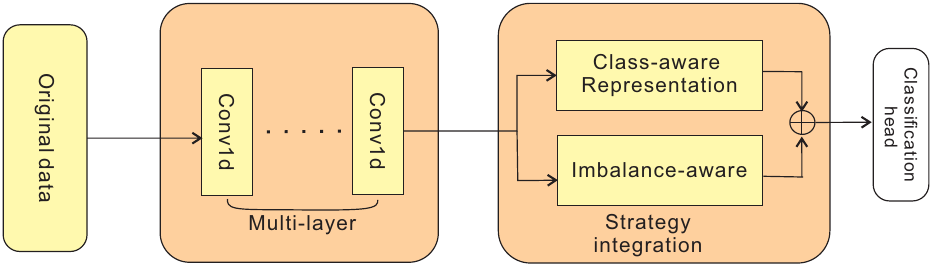}
    \caption{Workflow of DSTS. Eyemovement data is processed through convolutional layers, integrating class-aware and imbalance-aware mechanisms, before passing the features to the classifier for prediction.}
    \label{fig:placeholder}
\end{figure}

\subsection{Discrete and Short-term Dependencies in Eye Movement}

Eye movement sequence, collected from children viewing different paradigms, provides valuable insights into cognitive and behavioral patterns.

Let $S = \{s_1, s_2, \dots,s_t, \dots, s_T\}$ be an eye movement sequence, where $s_t \in \mathbb{R}^d$ denotes the eye-tracking feature vector at time step $t$, and $d$ is the dimensionality of the feature space. Each $s_t$ at each time step encapsulates a variety of ocular signals, such as gaze coordinates, duration, movement dynamics, and physiological measurements, forming a Multivariate Time-Series (MTS) that reflects the complex nature of eye movement behavior.

The eye movement data is inherently discrete, with each fixation point $s_t$ representing a distinct gaze focus at time step $t$. These discrete events are interspersed with saccades, which introduce discontinuities in the sequence. Additionally, gaze behavior exhibits short-term dependencies, where consecutive fixation points are strongly correlated, reflecting localized focus on immediate visual stimuli.
However, these methods are sensitive to class imbalance in the data, as imbalanced datasets can lead to biased models that perform poorly on underrepresented classes, ultimately impacting the accuracy and generalization of the model.

\subsection{Architecture}\label{sebsec_ar}

Modern architectures like Transformer-based models (e.g., MPTSnet~\cite{mu2025mptsnet}, and so on) are often employed to capture long-range temporal dependencies, as they are designed to model complex, global dependencies across long sequences. However, the characteristics of eye movement data—its discrete structure and short-term temporal dependencies—pose challenges for such models. In particular, eye-movement sequences consist of fixation points that remain temporally stable, yet the transitions between them (saccades) occur almost instantaneously, producing abrupt, non-smooth discontinuities that make the sequence inherently piecewise-discrete. While adjacent fixation points exhibit strong local correlations, these informative dependencies are largely restricted to short neighborhoods surrounding fixation–saccade events. Moreover, the behavioral cues relevant for ASD/TD discrimination primarily emerge from these localized patterns rather than from global temporal interactions. 

To address these challenges, we opt for the simplest yet effective approach: a Convolutional Neural Network (CNN). CNNs are well-suited for capturing local patterns within fixed-length windows, making them effective for modeling short-term dependencies between adjacent fixation points. Through convolutional filters, CNNs learn hierarchical features that capture the immediate gaze dynamics, efficiently modeling the short-term temporal correlations in eye movement data.

For an eye movement sequence $\mathbf{S}$, convolutional layers are employed to extract local patterns in the time-series data. Each convolutional layer applies a 1D convolution with a kernel to the input sequence, followed by batch normalization and ReLU activation to introduce non-linearity:

\begin{equation}
    \mathbf{z}_l = \text{ReLU}\left( \text{BatchNorm}_l\left( \text{Conv1d}_l(\mathbf{z}_{l-1}) \right) \right),
\end{equation}

where $\text{Conv1d}_l(\mathbf{z}_{l-1}) $ represents the $l$-th 1D convolution operation applied to the input sequence $\mathbf{S} $. The batch normalization step normalizes the feature maps, ReLU activation  help the model learn more complex patterns in the eye movement..

After passing through multiple convolutional layers, the resulting feature map $ \mathbf{z}_n $ captures hierarchical, local temporal patterns in the eye movement data. Max pooling over the time dimension is applied to obtain a fixed-length representation, known as the $\mathbf{e}$:

\begin{equation}
    \mathbf{e} = \text{MaxPool}(\mathbf{z}).
\end{equation}

After obtaining the embedding through the max pooling operation, the fixed-length representation is passed to a fully connected layer, which can be formulated as:

\begin{equation}
    y_{\text{pred}} = \text{Fc}(\mathbf{e}),
\end{equation}
where $ y_{\text{pred}} \in \mathbb{R}^c $ represents the predicted class scores, and $ C $ is the number of classes. The fully connected layer $ \text{Fc} $ acts as a classifier, applying a linear transformation to the embedding $e$ to obtain the prediction.

\subsection{Class-aware Representation Mechanism}

To further enhance the learned feature representations, we introduce the Class-Aware Representation Mechanism (CaR). This mechanism optimizes the feature space by ensuring that similar samples (e.g., gaze patterns from the same subject or similar behaviors) are placed closer together, while dissimilar samples (e.g., gaze patterns from different subjects or abnormal behaviors) are pushed further apart. By doing so, the model becomes more sensitive to subtle differences between eye movement patterns, improving its ability to distinguish between classes.
Mathematically, Multi-Similarity Loss (MS Loss) is introduced, which adjusts the feature space by minimizing the distance between similar samples and maximizing the distance between dissimilar ones, thereby optimizing the Class-aware Representation. The Loss is computed as follows:

\begin{equation}
\begin{split}
\mathcal{L}_{CaR} =& \frac{1}{n} \sum_{i=1}^{n} \left\{ \frac{1}{\alpha} \log \left[ 1 + \sum_{k \in P_i} e^{-\alpha(S_{ik} - \lambda)} \right] \right. +\\ 
&\left. \frac{1}{\beta} \log \left[ 1 + \sum_{k \in N_i} e^{\beta(S_{ik} - \lambda)} \right] \right\},
\end{split}
\end{equation}
where the terms with parameters $\alpha$ and $\beta$ control the margin for similar and dissimilar pairs. $\alpha$ regulates the contribution of positive pairs, while $\beta$ influences the negative pairs. $S_{ik}$ denotes the similarity of the $i$-th and $k$-th embedding of samples. $P_i$ is a positive sample set corresponding to the $i$-th sample, containing samples from the same class, and $N_i$ refers to the negative sample set.

\begin{table}[!t]
\caption{Dataset composition: The table summarizes the number of participants in the dataset, categorized by Autism Spectrum Disorder (ASD) and Typically Developing (TD) individuals.}
    \centering
    \begin{tabular}{l|c|c|c|c}
    \toprule
         Index&Dataset& Participants & ASD & TD\\
    \midrule
         1&Speaking& 1390 & 966&422\\
         2&Walking1&1356&939&417\\
         3&Walking2&1380&957&423\\
         4&Helicopter&1375&955&420\\
         5&Baby&1348&939&409\\
         6&Tablet&1372&954&418\\
         7&Attention&1349&939&410\\
         8&Sad&1277&882&395\\
    \bottomrule
    \end{tabular}
    \label{tb:data}
\end{table}

\begin{table*}[!t]
\caption{Recognition performance w.r.t. ACC and F1. \textbf{\textcolor{orange}{Orange}} and \textbf{\textcolor{gray}{Gray}} represent the best and second-best results, respectively.}
\label{tb:result}
\centering

\begin{tabular}{l|l|cccccccc}
\toprule
Method &Metrics& Speaking & Walking1 & Walking2 & Helicopter & Baby & Tablet & Attention & Sad \\
\midrule

Informer&ACC & 0.755 & 0.673 & 0.681 & 0.731 & 0.674 & 0.760 & 0.678 & 0.648 \\
~\cite{zhou2021informer} AAAI'21&F1& 0.722 & 0.649 & 0.649 & 0.679 & 0.674 & 0.716 & 0.651 & 0.630 \\
\midrule
Crossformer&ACC & 0.701 & 0.702 & 0.746 & 0.731 & 0.711 & 0.727 & 0.667 & 0.652 \\
~\cite{zhang2023crossformer} ICLR'23&F1 & 0.645 & 0.673 & 0.718 & 0.688 & 0.668 & 0.704 & 0.641 & 0.614 \\
\midrule
Detach-Rocket & ACC&\cellcolor{gray!20}0.781 & \cellcolor{gray!20}0.790 & 0.783 & \cellcolor{gray!20}0.771 & 0.793 & \cellcolor{gray!20}0.800 & 0.767 & \cellcolor{gray!20}0.766 \\
~\cite{uribarri2024detach} DMKD'24&F1 &\cellcolor{gray!20}0.849 & \cellcolor{gray!20}0.858 & 0.850 & 0.848 & 0.850 & \cellcolor{gray!20}0.865 & 0.840 & \cellcolor{gray!20}0.840 \\
\midrule

EmMixformer&ACC & 0.745 & 0.761 & \cellcolor{gray!20}0.790 & 0.756 & \cellcolor{gray!20}0.807 & 0.778 & \cellcolor{gray!20}0.774 & 0.734 \\
~\cite{qin2025emmixformer}'25&F1 & 0.839 & 0.828 & \cellcolor{gray!20}0.856 & \cellcolor{gray!20}0.860 & \cellcolor{gray!20}0.860 & 0.841 & \cellcolor{gray!20}0.844 & 0.820 \\
\midrule
MPTSNet&ACC & 0.691 & 0.669 & 0.721 & 0.735 & 0.748 & 0.687 & 0.663 & 0.656 \\
~\cite{mu2025mptsnet} AAAI'25&F1 & 0.817 & 0.814 & 0.737 & 0.817 & 0.817 & 0.780 & 0.752 & 0.780 \\
\midrule
DSTS  & ACC&\cellcolor{orange!40}0.813 & \cellcolor{orange!40}0.816 & \cellcolor{orange!40}0.804 & \cellcolor{orange!40}0.818 & \cellcolor{orange!40}0.819 & \cellcolor{orange!40}0.833 & \cellcolor{orange!40}0.789 & \cellcolor{orange!40}0.781 \\
(ours)&F1 & \cellcolor{orange!40}0.874 & \cellcolor{orange!40}0.872 & \cellcolor{orange!40}0.859 & \cellcolor{orange!40}0.871 & \cellcolor{orange!40}0.870 & \cellcolor{orange!40}0.878 & \cellcolor{orange!40}0.850 & \cellcolor{orange!40}0.845 \\

\bottomrule
\end{tabular}

\end{table*}

\begin{figure}
    \centering
    \includegraphics[width=1\linewidth]{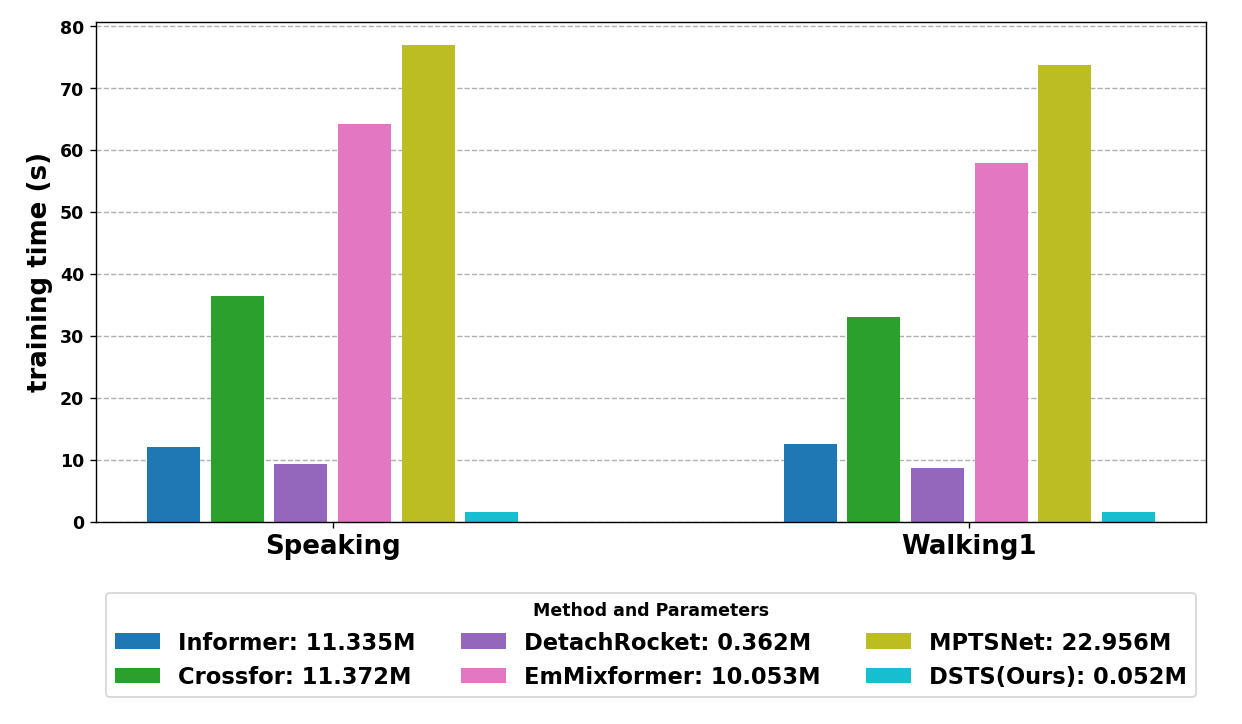}
    \caption{Comparison of time consumption and parameter count across methods on the Speaking and Walking1 datasets.}
    \label{fig:time}
\end{figure}

\begin{table*}[!t]
\caption{ACC performance of ablated DSTS variants formed by successively ablating the core components: MS Loss and WCE Loss. \textbf{\textcolor{orange}{Orange}} and \textbf{\textcolor{gray}{Gray}} represent the best and second-best results, respectively.}
\label{tb:ab_acc}
\centering
\resizebox{2\columnwidth}{!}{
\begin{tabular}{cc|c|cccccccc|c}
\toprule
\multicolumn{2}{c|}{DSTS Componets}& \multirow{2}{*}{Variants} & \multicolumn{8}{c|}{Datasets} & Average \\
\cmidrule(l{-0.15em}r){1-2}
\cmidrule(lr){4-11} 

CaR & Ia & &Speaking & Walking1 & Walking2 & Helicopter & Baby & Tablet & Attention & Sad &Rank\\

\midrule

$\checkmark$&$\checkmark$  & DSTS &
\cellcolor{orange!40}0.813 &  
\cellcolor{orange!40}0.816 &  
\cellcolor{orange!40}0.804 &  
\cellcolor{orange!40}0.818 & 
\cellcolor{gray!20}0.819 &  
\cellcolor{orange!40}0.833 &  
\cellcolor{orange!40}0.789 &  
\cellcolor{orange!40}0.781 &  
1.13  \\ 

&$\checkmark$  & ${I}$ &
0.789&
0.794&
\cellcolor{gray!20}0.797&
\cellcolor{gray!20}0.815&
0.815&
0.811&
\cellcolor{gray!20}0.789&
0.773 &

2.50  \\ 
$\checkmark$&  & ${II}$ &
\cellcolor{gray!20}0.799&
\cellcolor{gray!20}0.805&
\cellcolor{gray!20}0.797&
\cellcolor{gray!20}0.815&
\cellcolor{orange!40}0.822&
\cellcolor{gray!20}0.818&
\cellcolor{gray!20}0.789&
\cellcolor{gray!20}0.777&
1.75\\
\bottomrule

\end{tabular}}
\end{table*}

\begin{table*}[!t]
\caption{F1 performance of ablated DSTS variants formed by successively ablating the core components: MS Loss and WCE Loss. \textbf{\textcolor{orange}{Orange}} and \textbf{\textcolor{gray}{Gray}} represent the best and second-best results, respectively.}
\label{tb:ab_f1}
\centering
\resizebox{2\columnwidth}{!}{
\begin{tabular}{cc|c|cccccccc|c}
\toprule
\multicolumn{2}{c|}{DSTS Componets}& \multirow{2}{*}{Variants} & \multicolumn{8}{c|}{Datasets} & Average \\
\cmidrule(l{-0.15em}r){1-2}
\cmidrule(lr){4-11} 

CaR & Ia & &Speaking & Walking1 & Walking2 & Helicopter & Baby & Tablet & Attention & Sad &Rank\\

\midrule

$\checkmark$&$\checkmark$  & DSTS &
\cellcolor{orange!40}0.874&
\cellcolor{orange!40}0.872&
\cellcolor{gray!20}0.859&
\cellcolor{orange!40}0.871&
\cellcolor{gray!20}0.870&
\cellcolor{orange!40}0.878&
\cellcolor{gray!20}0.850&
\cellcolor{gray!20}0.845&
 
1.50  \\ 

&$\checkmark$  & ${I}$ &
0.855&
0.854&
0.856&
\cellcolor{gray!20}0.868&
0.866&
\cellcolor{gray!20}0.871&
\cellcolor{orange!40}0.854&
\cellcolor{orange!40}0.849&
2.25  \\ 

$\checkmark$&  & ${II}$ &
\cellcolor{gray!20}0.863&
\cellcolor{gray!20}0.865&
\cellcolor{orange!40}0.861&
0.867&
\cellcolor{orange!40}0.874&
0.869&
0.845&
0.843&
2.25\\
\bottomrule

\end{tabular}}
\end{table*}

\subsection{Imbalance-aware Mechanism}

Real-world medical data often exhibits class imbalance, where certain categories are underrepresented compared to others.
To address this issue, an Imbalance-aware mechanism (Ia) is employed. This mechanism adjusts the contribution of each class by assigning a weight based on its frequency in the training data, thereby mitigating the effects of imbalance and enhancing the model's performance on underrepresented classes. Specifically, samples from minority classes are assigned larger weights, so that their misclassification incurs a higher penalty during optimization. In this way, the training objective explicitly counteracts the bias toward majority classes and encourages more balanced decision boundaries.
This mechanism is implemented through the use of weighted cross-entropy Loss, computed as:

\begin{equation}
    \mathcal{L}_{\text{Ia}} = -\sum_{i=1}^{c} w_{y_i} \cdot p_{y_i}\log(p_{y_i}),
\end{equation}
where $ w_{y_i} $ is the weight for class $ y_i $, and $ p_{y_i} $ is the predicted probability for class $ y_i $. The weights $ w_{y_i} $ are computed based on the inverse frequency of classes, ensuring that classes with fewer samples contribute more to the Loss.

CaR and Ia are combined to enhance the model's ability to learn discriminative features from imbalanced data, preserving its capacity to classify subtle patterns in eye movement behavior. The overall loss is expressed as a weighted combination of the two components, formulated as:
\begin{equation}
    \mathcal{L}_{total} = \lambda \cdot \mathcal{L}_{CaR} + (1 - \lambda) \cdot \mathcal{L}_{Ia},
\end{equation}
where $\lambda$ controls the contribution of the CaR and Ia.

\section{Experiment}

\subsection{Experimental Setup.}
\textbf{Model Configuration.} Eye movement data is processed through a two-layer CNN as the embedding layer, integrating Class-aware Representation (CaR) and Imbalance-aware (Ia) mechanisms in DSTS. The $\lambda$ value for controlling the contribution of CaR and Ia is set to 0.25 across all datasets. The model is optimized using Adam optimizer.
\textbf{Five counterparts are compared:} Informer~\cite{zhou2021informer}, CrossFormer~\cite{zhang2023crossformer}, MPTSNet~\cite{mu2025mptsnet} are time-series analysis models. As for eye movement data analysis, EmMixformer~\cite{qin2025emmixformer} is proposed for Biometric identification, and Detach-Rocket~\cite{uribarri2024detach} is used to recognize Parkinson's disease. F1-score~\cite{f1} and Accuracy (ACC)~\cite{acc} are utilized for detection performance evaluation. All experiments are implemented on an i7-13700HX CPU and a 4060 Laptop GPU.

\textbf{Dataset.}
We used a set of eight eye movement datasets, namely: Speaking, Walking1, Walking2, Helicopter, Baby, Tablet, Attention, and Sad, collected by the Shenzhen Maternity and Child Healthcare Hospital. These datasets correspond to different paradigms, each capturing distinct aspects of eye movement behavior. The specific purpose of this data collection was to distinguish children with Autism Spectrum Disorder (ASD) from their Typically Developing (TD) peers, all collected data can be utilized to assist healthcare professionals in the diagnosis of ASD in children. The total number of samples, as well as the respective sample counts for ASD and TD, are listed in the Table~\ref{tb:data}. Notably, the datasets exhibit class imbalance, with a higher number of children diagnosed with ASD compared to TD children. 

\subsection{Performance Evaluation.}
The detection performance of different algorithms is investigated to statistically analyze the superiority of DSTS.
\textbf{Detection performance of different methods} is compared in Table~\ref{tb:result} w.r.t. ACC and F1-score, respectively. The best and second-best results on each dataset are highlighted in \textbf{\textcolor{orange}{Orange}} and \textbf{\textcolor{gray}{Gray}}, respectively. The observations include the following two aspects: (1) Overall, DSTS performs best on almost all datasets and significantly outperforms the second-best method on most datasets, indicating its superiority in detection. (2) The F1-score and accuracy (ACC) of DSTS on the Attention and Walking2 datasets are marginally higher than those of the second-best method. However, the second-best method, EmMixformer, performs worse than DSTS on the other datasets, which indicates the robustness of DSTS.
\textbf{Comparison of time consumption and parameter scale across different methods} are shown in Fig.~\ref{fig:time} for the Speaking and Walking1 datasets. DSTS consumes the least time and the fewest parameters compared to its counterparts, achieving the best performance across all datasets, thereby validating its efficiency and effectiveness in real-world applications.

\subsection{Ablation Study.}
To further validate the effectiveness of different modules, ablation studies are conducted on the two Loss functions used: CaR and Ia. Specifically, we removed CaR to create ${I}$ and removed Ia to create ${II}$. The results are presented in Table~\ref{tb:ab_acc} and Table~\ref{tb:ab_f1}.
The experimental results demonstrate that the proposed method achieves the best average performance. It attains the highest ACC on all datasets except the Baby dataset. In terms of F1-score, it achieves the best or second-best performance across all datasets, thereby validating the contribution of each module.

\section{Concluding Remarks}

In this study, we proposed a discrete short-term sequential (DSTS) modeling framework for early Autism Spectrum Disorder (ASD) detection using eye movement data. By integrating CNNs with Class-aware Representation and Imbalance-aware Mechanisms, our method captures complex eye movement patterns to distinguish ASD from typically developing (TD) individuals. Experimental results show that DSTS outperforms both traditional and Transformer-based models, demonstrating its efficiency and practical applicability.

Future work should investigate the reasons behind the limited performance of complex models, such as Transformers, on eye movement data. In addition, it is essential to explore the development of a general framework that enables models to adaptively learn from eye movement data across multiple paradigms and stimuli.

\bibliography{refs.bib} 
 
\bibliographystyle{IEEEtran}

\end{document}